\documentclass[12pt, final]{l4dc2022}

\usepackage{times}
\usepackage{wrapfig}
\usepackage[font={footnotesize}]{caption}

\long\def\ignorethis#1{}

\newcommand{\eg}{e.g.\ }
\newcommand{\ie}{i.e.\ }

\newcommand{\vc}[1]{\ensuremath{\mathbf{#1}}}

\title[Data-Augmented Contact Model for Rigid Body Simulation]{\bf{Data-Augmented Contact Model for Rigid Body Simulation}}

\author{%
 \Name{Yifeng Jiang} \Email{yifengj@stanford.edu}\\
 \addr Stanford University, California, United States
 \AND
 \Name{Jiazheng Sun} \Email{jiazhengsun96@gmail.com}\\
 \addr Amazon.com, Inc., Washington, United States 
 \AND
 \Name{C. Karen Liu} \Email{karenliu@cs.stanford.edu}\\
 \addr Stanford University, California, United States
}

\begin{document}

\maketitle

\begin{abstract}%
 Accurately modeling contact behaviors for real-world, near-rigid materials remains a grand challenge for existing rigid-body physics simulators. This paper introduces a \emph{data-augmented} contact model that incorporates analytical solutions with observed data to predict the 3D contact impulse which could result in rigid bodies bouncing, sliding or spinning in all directions. Our method enhances the expressiveness of the standard Coulomb contact model by learning the contact behaviors from the observed data, while preserving the fundamental contact constraints whenever possible. For example, a classifier is trained to approximate the transitions between static and dynamic frictions, while non-penetration constraint during collision is enforced analytically. Our method computes the aggregated effect of contact for the entire rigid body, instead of predicting the contact force for each contact point individually, maintaining same simulation speed as the number of contact points increases for detailed geometries. Supplemental video: \url{https://shorturl.at/eilwX}
\end{abstract}




\section{Introduction}

The use of virtual simulations has been crucial to the recent progress in building learning-based robot controllers. With recent works showing that improving the rendering realism of simulated perception \cite{rao2020rl} or the accuracy of robot motor dynamics \cite{hwangbo2019learning} can improve the performance of learned controllers, more accurate modeling of the contact dynamics would also help the robot better understand and interact with the environment it resides. Most recent robot controllers are still developed from off-the-shelf physics simulators which use the idealized Coulomb friction model, an empirical construct to approximate the changes between two physical regimes (static friction vs dynamic friction). However, the Coulomb model assumes linear relationship between normal force and frictional force and uses a single friction coefficient to represent an isometric friction cone. The computation of contact force also involves approximation and arbitrary decisions. For example, many existing simulators formulate a Linear Complementarity Program (LCP), which solutions are not unique except for the frictionless case \cite{brogliato1999nonsmooth}. Depending on the initial guesses and numerical methods used for solving the LCP problem (\eg Lemke method \cite{Lemke} vs Gauss-Seidel algorithm \cite{GS}), the resulting contact forces can be drastically different. These existing issues suggest that a computational contact model grounded by observed data can be a desired alternative.




While recent work has shown that physical phenomena can be learned from data and approximated by neural networks, precisely enforcing constraints, such as contacts, remains difficult for these function approximators learned in an end-to-end fashion. Consider a box resting on a table. If the contact force is slightly larger or smaller than the gravitational force, we will start to see the box rattling or sinking into the table without any external force. Such \emph{categorically} incorrect prediction of physics outcomes are likely to have negative impacts on the development of control policies.

This paper introduces a \emph{data-augmented} contact model combining analytical solutions and empirical data collected for a particular scenario (\eg a specific robot foot colliding a specific surface), such that the simulated trajectories better match the observed data. 

Our approach is built on two key insights. First, we  utilize analytical solutions from first principles whenever possible and only resort to data-driven approach when the phenomenon is less well understood. For example, the contact force that prevents the objects from interpenetrating is enforced by equations instead of learned from the data. In contrast, we rely on observed data to model and validate the less-understood static friction boundary. To this end, we propose to decompose the contact problem to two steps: predicting the next contact state (\ie static, dynamic, or detach) and determining contact forces. We solve the first step by learning a classifier from the observed data and the second step by a combination of learning a regressor and solving constrained systems. Second, we propose to compute the aggregated effect of contact at the rigid-body level, instead of predicting the contact force at each individual contact point. This decision removes the concern of simulation speed as the number of contact points increases for real-world objects with detailed geometries.


Our algorithm also applies to articulated systems, distinguishing our framework from most data-driven contact models. Once a rigid body's contact model is trained, we can simply connect the rigid body to an articulated system without retraining the contact model. This implies that the collision data for learning the contact can be collected using replicas of the disassembled end-effector, without putting the entire robotic system at risk.

As our work focuses on a full 3D scenario in which the object can bounce, slide, or spin in all directions, existing contact data such as MIT Push \cite{yu2016more} are not diverse enough for our evaluation. Instead, we learn the data-driven models from synthesised collision data in simulations. Evaluations show that our data-augmented model matches ground-truth contact behaviors of both single and articulated rigid body systems, and better than a purely statistical model.

\section{Related Work}

Contact and friction is a common but extremely complex phenomenon, which continuously fascinates generations of scientists and engineers. Since Coulomb and Amontons in the 18th century made the distinction between static and dynamic frictions, inadequacies and controversies of Coulomb's law have been extensively studied \cite{Elena}. For example, \cite{oden1985models} summarized that friction could depend on normal force or stress, normal separation distance, slip displacement, slip velocity, time of stationary contact, slip history, and vibrations. As such, many empirical models that substitute Coulomb's law have been proposed and are summarized in \cite{olsson1998friction}. There is no one model that is more accurate than the others in all scenarios. \cite{goyal1989limit} proposed the concept of limit surface to enclose all possible friction forces on an object during planer sliding. Our 3D model also works on the rigid-body instead of the point-contact level, and loosens the assumptions that friction must oppose the direction of motion and that friction is isotropic.

In computer animation and robotics, the simplest Coulomb's law and the point contact representation are often used to approximate contact physics for visualizing or evaluating robotic algorithms. Existing rigid-body simulators often solve a Linear Complementarity Problem (LCP) by enforcing unilateral constraints and complementarity between relative velocities and contact forces \cite{baraff1989analytical, stewart2000implicit, anitescu2002time}. In contrast, \cite{todorov2014convex} relaxed contact constraints to solve a convex optimization problem at each time step. 
With the emergence of simulation-based controller learning, developing data-driven physics simulators capable of predicting the real world has been actively studied. \cite{chang2016compositional,byravan2017se3,lerer2016learning} trained neural networks to learn physical intuitions from vision perception, \cite{zhou2016convex} built a data-efficient model for the limit surface of an object during planar sliding, \cite{bauza2017probabilistic} built a probabilistic model for planer sliding that takes into account the stochasticity of frictional forces, and \cite{fazeli2017learning} proposed either to train a purely data-driven model, or to use data to learn the optimal parameters of an analytical model for planer impact. Our work is instead a data-augmented one, incorporating observed data while preserving fundamental contact constraints analytically. Contact detection (not handling) could also be replaced with learnable components via learning contact Jacobian \cite{pfrommer2020contactnets} or differentiable shape parameterization \cite{strecke2021_diffsdfsim}. System identification \cite{chebotar2019closing} serves as another approach to improve simulation fidelity without requiring new simulation models, with recent works demonstrating promising results by utilizing gradient-free reinforcement learning \cite{jeong2019modelling, 9561731}, or making simulators differentiable \cite{9560935, 9363565, 10.3389/fnbot.2019.00006, gradsim}.

The advantages of integrating neural networks with analytical models have been demonstrated recently. \cite{ajay2018augmenting} introduced a method that trains a recurrent neural network to predict the deviation between real-world and simulated contact trajectories, extended in \cite{fazeli2020long} for better long-term accuracy. \cite{kloss2018} proposed to combine a neural network for perception with a physics model for prediction, and \cite{pizzuto2021physics} added a non-penetration loss to augment the data loss. Instead of treating the physics engine as a black box and correcting its output, our work directly improves the full 3D contact handling using learned function approximators. We also show that our learned contact models can be reused in new articulated systems without retraining.

\section{Method}

We propose a method to predict the contact impulse between a specific pair of near-rigid objects. An ideal contact model in a physics simulator should at least guarantee the following properties:
\begin{enumerate}
\vspace{-0.5pc}
\item{Non-penetration:} The geometries of the objects in contact should not overlap.
\vspace{-0.5pc}
\item{Repulsive force:} The contact force should only push the objects away instead of pulling them together.
\vspace{-0.5pc}
\item{Workless condition:} The contact force becomes zero at the instance when the bodies begin to separate.
\vspace{-0.5pc}
\item{Two friction regimes:} There is an unsmooth switch between static and dynamic friction forces, depending on the materials of the objects and other factors.
\vspace{-0.5pc}
\item{Dynamic friction model:} The dynamic friction force depends on the normal contact force, the relative velocity of the objects, and other factors.
\vspace{-0.5pc}
\end{enumerate}

Among these five properties, (4) and (5) depend on empirical models because their mechanics are not well understood. As such, our method will use a data-driven approach to achieve (4) and (5), while maintaining the analytical solution that satisfies (1)-(3). 

\subsection{Assumptions}
We address a perfectly inelastic (\ie the coefficient of restitution is zero), non-adhesive contact phenomenon between two near-rigid objects, which is a common scenario in robotic applications. We assume both objects have convex shapes and that one of them is stationary. The deformation during the collision is negligible comparing to the overall rigid body motion. We expect the range of collision impulses encountered at test time to lie within the range used for training. For collision between articulated rigid body systems, we assume that the distal link is the only part of the system that is in contact.

We compute the aggregated effect of contact for the entire rigid body, instead of using point-contact representation, which increases computational complexity and leads to issues with over-parameterization. Our method represents the contact geometry as a patch with non-zero area on the surface of the object in contact, denoted as $\mathcal{P}$. We assume that the normal direction of the contact patch is well-defined.


\subsection{Single rigid body}

\begin{wrapfigure}[21]{l}{2.6in}
\vspace{-0.5pc}
\begin{algorithm2e}[H]
  \LinesNumbered
  \SetKwInOut{Data}{Input}
  \SetKwInOut{Result}{Output}
  \Data{$\vc{q}_{t}, \dot{\vc{q}}_{t},\boldsymbol{\tau}, \mathcal{P}$}
  \Result{$\vc{p}$}
  $\dot{\vc{q}}^{(1)} = \dot{\boldsymbol{q}}_{t} + h \vc{M}^{-1}\boldsymbol{\tau}$\;
  $c \leftarrow C(\vc{q}_{t}, \dot{\vc{q}}^{(1)})$\;
  \uIf{$c ==$"static"}{
  AddPositionConstraint($\mathcal{P}$)\;
  $\vc{p} \leftarrow$ SolveConstraint($\vc{q}_{t}, \dot{\vc{q}}^{(1)}$)\;
  RemovePositionConstraint($\mathcal{P}$)\;
  }
  \uElseIf{$c ==$"dynamic"}{
  $\vc{p}_f \leftarrow R(\vc{q}_{t}, \dot{\vc{q}}^{(1)})$\;
  $\dot{\vc{q}}^{(2)} = \dot{\vc{q}}^{(1)} + \vc{M}^{-1}\vc{T}_f \vc{p}_f$\;
  $\vc{p}_n \leftarrow$ FrictionlessLCP($\vc{q}_{t}, \dot{\vc{q}}^{(2)}$)\;
  $\vc{p} = \vc{T}_n \vc{p}_n + \vc{T}_f \vc{p}_f$\;
  }
  \Else{
  $\vc{p} = \vc{0}$\;
  }
  \Return{$\vc{p}$}
\caption{Single body contact model: $H$}
\label{algo_single_body}
\end{algorithm2e}
\end{wrapfigure}

The contact computation in physics engines typically consists of two separate processes, contact detection that identifies the contact location on the surface of the object, and contact handling that calculates the contact force such that the contact constraints and the equations of motion are satisfied. Using a standard contact detector, $D(\vc{q}_t)$, we can compute a contact patch $\mathcal{P}$, represented by the convex hull of contact points, given shape and current position of the rigid body $\vc{q}_t \in \mathbb{R}^6$ expressed in the generalized coordinates.

The main challenge of contact computation lies in contact handling. In its most general form, a contact handling routine can be expressed as a function $\vc{p} = H(\vc{q}_t, \dot{\vc{q}}_t, \boldsymbol{\tau}, \mathcal{P})$, which maps the pre-contact state (\ie $\vc{q}_t$ and $\dot{\vc{q}}_t$), applied forces $\boldsymbol{\tau}$, and contact patch $\mathcal{P}$ to the 6D contact impulse, $\vc{p}$. Precisely, $\vc{p}$ is the integrated pressure on the contacting surface over the entire contact patch $\mathcal{P}$ and over the time step interval $h$. During the collision process, the contact pressure might not be constant, but its aggregated effect in one time step is equivalent to the impulse $\vc{p}$, which can be used to integrate the state forward to the next time step by
\begin{eqnarray}
\dot{\vc{q}}_{t+1} = \dot{\vc{q}}_t + \vc{M}(\vc{q}_t)^{-1}(h \boldsymbol{\tau} + \vc{p}), \\
\vc{q}_{t+1} = \mathrm{Integrate}(\vc{q}_{t}, \dot{\vc{q}}_{t+1}),
\end{eqnarray}
where $\vc{M}(\vc{q}_{t})$ is the generalized mass matrix of the rigid body. At every time step, if $D$ detects a non-empty $\mathcal{P}$, we invoke the contact handler $H$, described in Algorithm \ref{algo_single_body}. 

We first update the current velocity $\dot{\vc{q}}_t$ to an intermediate velocity by explicitly integrating the applied force $\boldsymbol{\tau}$: $\dot{\vc{q}}^{(1)} = \dot{\vc{q}}_t + h \vc{M}^{-1} \boldsymbol{\tau}$. Directly training a regressor to predict the contact impulse $\vc{p}$ is likely to violate Properties (1)-(3), as they require precise satisfaction of constraints. Instead, we use observed data to train a classifier $C(\vc{q}_t, \dot{\vc{q}}^{(1)})$ that predicts one of the following outcomes for $\mathcal{P}$: \emph{static}, \emph{dynamic}, or \emph{detach}. Based on the predicted outcome, we will calculate the contact impulse differently in order to satisfy (1)-(3).

\begin{wrapfigure}{r}{3.5in}
\centering
  \includegraphics[width=0.5\textwidth]{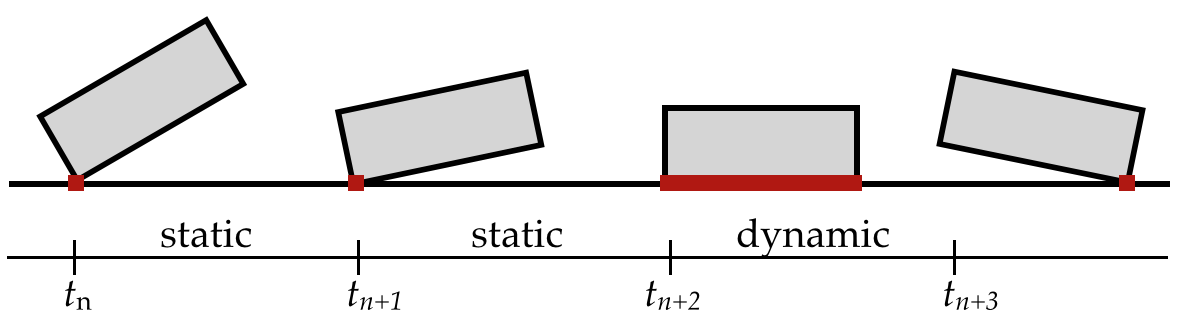}
  \caption{Illustration of "static" and "dynamic" cases. The classifier predicts "static" at $t_n$ and $t_{n+1}$. We analytically solve a contact impulse to ensure the contact patch (shown in red) has zero velocity at the end of $t_n$ and $t_{n+1}$. The classifier predicts "dynamic" at $t_{n+2}$ and expects the contact patch to change at the end of $t_{n+2}$.}
    \label{fig:method-contactstates}
\end{wrapfigure}
  


\paragraph{Static case}
The static case indicates that $\mathcal{P}$ will remain in the same position and orientation at the beginning of next time step, but the rigid body is not necessarily stationary. Fig. \ref{fig:method-contactstates} illustrates an example in which the rigid body is moving while $\mathcal{P}$ is static. The contact impulse in this case must ensure that $\mathcal{P}$ has zero velocity at the end of this time step. In physics engines, this is easily achieved by setting positional constraints at $\mathcal{P}$ and solve for the constraint impulse $\vc{p}$ that satisfies $\vc{v}_p = \vc{0}$, where $\vc{v}_p$ is the generalized velocity of $\mathcal{P}$ at the end of time step. Since $\vc{v}_p$ and $\vc{p}$ have the same effective degrees of freedom, the linear system $\vc{v}_p(\vc{p}) = \vc{0}$ has a unique solution. Therefore, if a state is classified as "static" by a highly accurate classifier $C$, the solution of $\vc{v}_p(\vc{p}) = \vc{0}$ will satisfy Properties (1)-(3). 



\paragraph{Dynamic case}
In the dynamic case, the contact patch $\mathcal{P}$ will change its position or orientation, or lose some area at the next time step (Fig. \ref{fig:method-contactstates}). To compute the motion of $\mathcal{P}$, we need to predict the contact impulse $\vc{p}$. We propose to train a regressor from the observed data because the idealized Coulomb friction model is limited when approximating the complex dynamic friction phenomena. However, directly using a regressor to predict $\vc{p}$ will still suffer from the same problem of failing to satisfy Properties (1)-(3) precisely. 

Instead, our algorithm first reparameterizes and decouples $\vc{p} = (p_x, p_y, p_z, m_x, m_y, m_z)^T$, where $p$ and $m$ indicate linear impulse and impulsive torque respectively, into a frictional impulse $\vc{p}_f = (p_x, p_z, m_y)^T$ and a normal impulse $\vc{p}_n = (p_y, \tilde{m}_x, \tilde{m}_z)^T$. We define $\tilde{m}_x$ and $\tilde{m}_z$ to be the impulsive torques induced by the normal linear impulse $p_y$. With this decoupling, in a perfectly inelastic case (\ie restitution is zero), given any $\vc{p}_f$, there exists one unique $\vc{p}_n$ such that Properties (1)-(3) are satisfied \footnote{The actual distribution of normal force over the contact patch is still undetermined. \cite{baraff1992dynamic}}. Therefore, we train a regressor to only predict $\vc{p}_f$ and analytically calculate the unique solution for $\vc{p}_n$ based on the predicted $\vc{p}_f$. Specifically, Algorithm \ref{algo_single_body} in the dynamic case first predicts $\vc{p}_f$ using the trained regressor $R(\vc{q}_t, \dot{\vc{q}}^{(1)})$ (Line 8) and then integrates $\vc{p}_f$ to obtain a second intermediate velocity: $\dot{\vc{q}}^{(2)} = \dot{\vc{q}}^{(1)} + \vc{M}^{-1}\vc{T}_f \vc{p}_f$ (Line 9). Here $\vc{T}_f \in \mathbb{R}^{6\times3}$ transforms $\vc{p}_f$ to the generalized coordinates. The unique solution for $\vc{p}_n$ can be solved by any routine that respects normal complementaries. Our algorithm uses a Danzig-like positive definite LCP solver for frictionless contacts \cite{COTTLE1968103} (Line 10). Finally, we combine the analytical $\vc{p}_n$ and predicted $\vc{p}_f$ to obtain $\vc{p}$ in generalized coordinates (Line 11).

If the regressor $R$ were perfectly accurate, the uniqueness of $\vc{p}_n$ ensures that the decoupling treatment in Algorithm \ref{algo_single_body} does not affect the true solution of the contact impulse $\vc{p}$. When $R$ is not perfectly accurate, the frictionless LCP (Line 10) serves as a corrective step on $\vc{p}$ that prioritizes the satisfaction of Properties (1)-(3) over Property (5).

\paragraph{Detach case}
In the detach case, $\mathcal{P}$ is predicted to have positive normal velocity and leave the surface at the next time step. We thus set $\vc{p} = \vc{0}$, ensuring that Property (3) is satisfied.

\textbf{Remarks:} Our method addresses Property (4) by learning a classifier from the observed data. Similarly, the regressor incorporates the observed data to address Property (5). Assuming that the classifier is highly accurate, the resultant contact impulse for the static and detach cases will closely match reality and satisfy Properties (1)-(3) exactly. The classifier will be less accurate near the decision boundary, which coincides with the poorly-understood region where transitions between different physical regimes occur. 

\subsection{Articulated rigid bodies}

\begin{wrapfigure}{r}{2.5in}
\centering
\vspace{1.5pc}
   \includegraphics[width=0.4\textwidth]{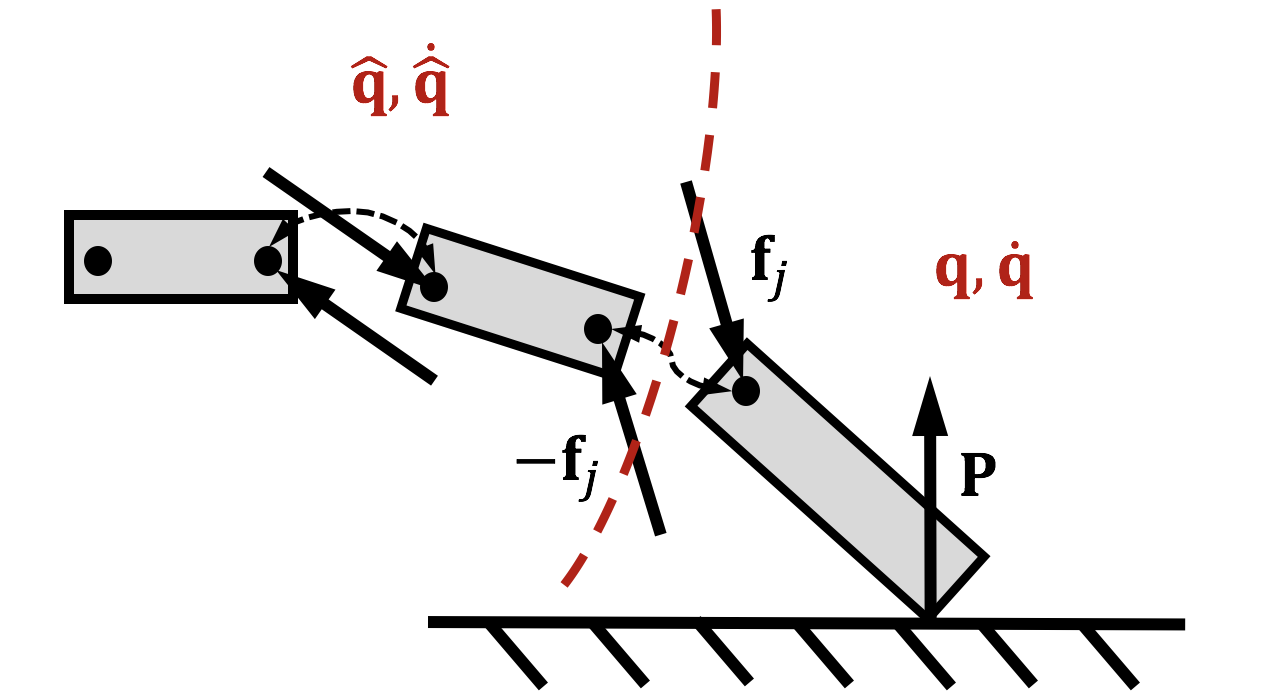}
  \caption{An articulated rigid body system which consists of a distal body whose state is $(\vc{q}, \dot{\vc{q}})$ and other bodies in the upstream system $(\hat{\vc{q}}, \dot{\hat{\vc{q}}})$.}
    \label{fig:multi-link-method}
    \vspace{1.5pc}
\end{wrapfigure}

Our method can be extended to articulated rigid body systems. We decompose the state of the system into the distal body $(\vc{q}_t, \dot{\vc{q}}_t)$ and all other bodies in the upstream system $(\hat{\vc{q}}_{t}, \dot{\hat{\vc{q}}}_{t})$ (Fig. \ref{fig:multi-link-method}). The joint force $\vc{f}_j$ transmitted between the distal body and the upstream system is unknown and must be solved \emph{simultaneously} with the contact impulse $\vc{p}$.

Algorithm \ref{algo_articulated} starts with expressing the velocity of the distal body at the next time step:
\begin{equation}
\dot{\vc{q}}_{t+1} = \dot{\vc{q}}_t + h \vc{M}^{-1}\boldsymbol{\tau} + h\vc{M}^{-1} \vc{J}^T\vc{f}_{j} + \vc{M}^{-1}\vc{p},
\end{equation}
where $\vc{M}(\vc{q}_{t})$ is the generalized mass matrix for the distal body and $\vc{J}(\vc{q}_{t})$ is the Jacobian transforming from the generalized coordinates of the rigid body to the Cartesian space at the joint. The Cartesian velocity at the joint at the next time step is then given by 
\begin{equation}
\vc{v}_{t+1} = \vc{J} \dot{\vc{q}}_t + h \vc{J}\vc{M}^{-1} \boldsymbol{\tau} + h \vc{J} \vc{M}^{-1} \vc{J}^T\vc{f}_{j} + \vc{J} \vc{M}^{-1}\vc{p}.
\end{equation}

Similarly, the velocity of the upstream system evaluated at the joint can be expressed as
\begin{equation}
\hat{\vc{v}}_{t+1} = \hat{\vc{J}} \dot{\hat{\vc{q}}}_t + h \hat{\vc{J}} \hat{\vc{M}}^{-1} \hat{\boldsymbol{\tau}} - h \hat{\vc{J}} \hat{\vc{M}}^{-1} \hat{\vc{J}}^T \vc{f}_{j},
\end{equation}
where $\hat{\vc{M}}$ and $\hat{\vc{J}}$ are the mass matrix and Jacobian for the upstream system. 

Since the joint constraint is satisfied at the beginning of the time step $t_0$, we only need to ensure that the velocity of the constraint is satisfied so that at $t_1$ the distal body and the upstream system still coincide at the joint. Therefore, we need to solve for a $\vc{f}_j$ such that $\vc{v}_{t+1} - \hat{\vc{v}}_{t+1} = \vc{0}$: 
\begin{equation}
G(\vc{f}_{j}) = \vc{v}_{t+1} - \hat{\vc{v}}_{t+1} = \vc{A}\vc{f}_{j} + \vc{J}\vc{M}^{-1} H(\vc{f}_{j}) + \vc{c} = \vc{0},
\label{functionG}
\end{equation}
where $\vc{A} = h \vc{J} \vc{M}^{-1} \vc{J}^T + h \hat{\vc{J}} \hat{\vc{M}}^{-1} \hat{\vc{J}}^T$ and $\vc{c} = \vc{J} \dot{\vc{q}}_t + h \vc{J} \vc{M}^{-1}\boldsymbol{\tau} - (\hat{\vc{J}} \dot{\hat{\vc{q}}}_t + h \hat{\vc{J}} \hat{\vc{M}}^{-1} \hat{\boldsymbol{\tau}})$ are constants in the equation given $\vc{q}_t$ and $\dot{\vc{q}}_t$. $H(\vc{f}_j)$ is a shorthand for $H(\vc{q}_t, \dot{\vc{q}}_t, \boldsymbol{\tau} + \vc{J}^T \vc{f}_{j}, \mathcal{P})$, which outputs $\vc{p}$ depending on $\vc{f}_j$.

\begin{wrapfigure}{l}{2.5in}
\centering
\begin{algorithm2e}[H]
  \LinesNumbered
  \SetKwInOut{Data}{Input}
  \SetKwInOut{Result}{Output}
  \Data{$(\vc{q}_{t}, \dot{\vc{q}}_{t},\boldsymbol{\tau}), (\hat{\vc{q}}_{t}, \dot{\hat{\vc{q}}}_{t},\hat{\boldsymbol{\tau}}), \mathcal{P}$}
  \Result{$\vc{q}_{t+1},\dot{\vc{q}}_{t+1}, \hat{\vc{q}}_{t+1}, \dot{\hat{\vc{q}}}_{t+1}$}
  
  Initialize $\vc{f}_{j}$ using Eq. \ref{initialGuess}\;
  Calculate $\vc{J}, \vc{M}, \hat{\vc{J}}, \hat{\vc{M}}$\;
  \While{solver not terminated}{
    Evaluate $G(\vc{f}_{j})$ using Eq. \ref{functionG}\;
    Update $\vc{f}_{j}$ according to Powell's method\;
  }
  $\vc{p} \leftarrow H(\vc{q}_{t}, \dot{\vc{q}}_{t}, \boldsymbol{\tau}+ \vc{J}^T \vc{f}_{j}, \mathcal{P})$\;
  $\dot{\vc{q}}_{t+1} = \dot{\vc{q}}_t + h\vc{M}^{-1}\boldsymbol{\tau} + h\vc{M}^{-1} \vc{J}^T \vc{f}_{j} + \vc{M}^{-1}\vc{p}$\;
  $\vc{q}_{t+1}$ = Integrate($\vc{q}_{t}, \dot{\vc{q}}_{t+1}$)\;
  $\dot{\hat{\vc{q}}}_{t+1} = \dot{\hat{\vc{q}}}_{t} + h \hat{\vc{M}}^{-1}\hat{\boldsymbol{\tau}} - h \hat{\vc{M}}^{-1} \hat{\vc{J}}^T \vc{f}_{j}$\;
  $\hat{\vc{q}}_{t+1}$ = Integrate($\hat{\vc{q}}_{t}, \dot{\hat{\vc{q}}}_{t+1}$)\;
  \Return{$\vc{q}_{t+1},\dot{\vc{q}}_{t+1}, \hat{\vc{q}}_{t+1}, \dot{\hat{\vc{q}}}_{t+1}$}
  
\caption{Contact solver for articulated rigid-body chain}
\label{algo_articulated}
\end{algorithm2e}
\end{wrapfigure}

We solve Eq. \ref{functionG} using Powell hybrid method \cite{powell1970hybrid}, which uses finite difference to approximate the Jacobian matrix and is less sensitive to the initial guess to the problem. Powell's method only requires a routine to evaluate $G(\vc{f}_j)$ and an initial guess. Using the heuristic that assumes $\hat{\vc{v}}_{t+1} = \vc{0}$, we compute the initial $\vc{f}_j$ by
\begin{equation}
\label{initialGuess}
\vc{f}_{j} = (h \hat{\vc{J}} \hat{\vc{M}}^{-1} \hat{\vc{J}}^T)^{-1}  \hat{\vc{J}} (\dot{\hat{\vc{q}}}_t + h \hat{\vc{M}}^{-1}\hat{\boldsymbol{\tau}}).
\end{equation}

In our experiment a solution can always be found at each time step with 0.5\% convergence tolerance. The number of evaluations of $G$ is often fewer than 10.


\subsection{Implementation}
A contact patch $\mathcal{P}$ in the real world will always be a 2D surface. However, when $\mathcal{P}$ degenerates to nearly an edge or a point, in practice, the dimension of the controllable space of the friction impulse will reduce. Since the dimension of $\mathcal{P}$ is available from the collision detector $D$, we utilize this information to improve learning accuracy by treating three types of $\mathcal{P}$ separately: a surface (2D), a line (1D), or a point (0D). For each type, we train a specific classifier and a regressor. Using separate neural networks allows the regressors to have different output dimensions according to the dimension of controllable space of the friction impulse. 

The same set of training data can be used to train the classifiers and the regressors. The data collection involves throwing objects to each other with different initial velocities. Since $\boldsymbol{\tau}$ is not part of the input of the learned models, we do not need to apply various $\boldsymbol{\tau}$ during training sample generation, greatly simplifying the data collection process. We record the entire trajectory for each throw and extract the state of every contact instance: $\vc{q}_{t}, \dot{\vc{q}}_{t}$. By evaluating $\dot{\vc{q}}_{t}$ at the patch, we can identify and label “static” and “dynamic” cases. To determine “detach” cases and to calculate the training output for the regressors, we need to recover the contact impulse $\vc{p}$ for each contact instance: $\vc{p} = \vc{M}(\dot{\vc{q}}_{t+1} - \dot{\vc{q}}_t) - h\vc{g}$, where $\vc{g}$ is the gravitational force. If $\vc{p}$ is near zero, we label this contact instance "detach". 

The range of initial velocities is chosen to cover the range of the anticipated collision impulses during testing. We found that the choice of rotation representation significantly affects the accuracy of the learned models. Our experiments show that representing the 3D orientation of the rigid body as a rotation matrix outperforms other representations of SO(3). We also found that  including two redundant features---the position of the center of $\mathcal{P}$ in the body frame and the velocity at the center of $\mathcal{P}$---reduces the errors of the regressors.


 \section{Evaluation}
We evaluated our data-augmented contact model on rigid bodies with different 3D shapes, a 3-linked articulated rigid body chain, and an object with anisotropic friction coefficient. To demonstrate the complexity of 3D collision, we also included one 2D example for comparison. The ground-truth data were collected in a simulated environment using a generic physics engine DART \cite{DART}.

\begin{wrapfigure}{r}{3.6in}
\centering
   \includegraphics[width=3.5in]{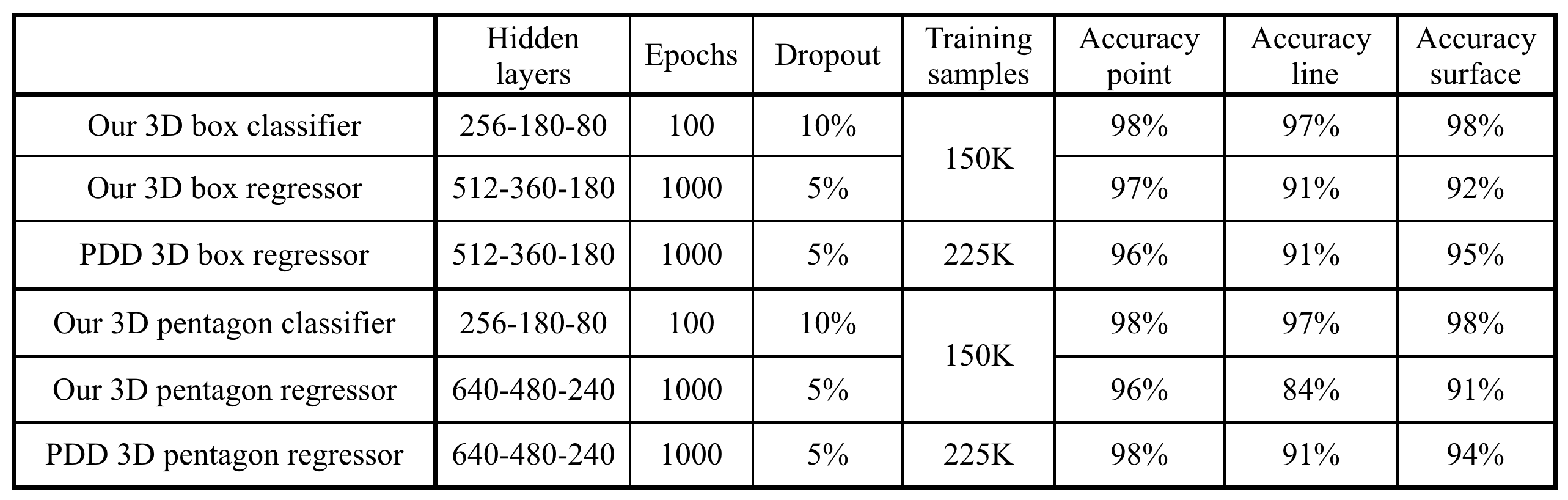}
  \caption{Hyper-parameters and testing accuracy.}
  \label{fig:params}
\end{wrapfigure}



Although our method is agnostic to the representation of classifiers or regressors, we used feed-froward neural networks for their expressiveness as function approximators. A standard cross-entropy error or MSE was used as the loss function. We used the same collision data to train classifiers and regressors. The hyper-parameters are shown in Fig. \ref{fig:params}.

\begin{wrapfigure}{r}{3.6in}
\centering
   \includegraphics[width=3.5in]{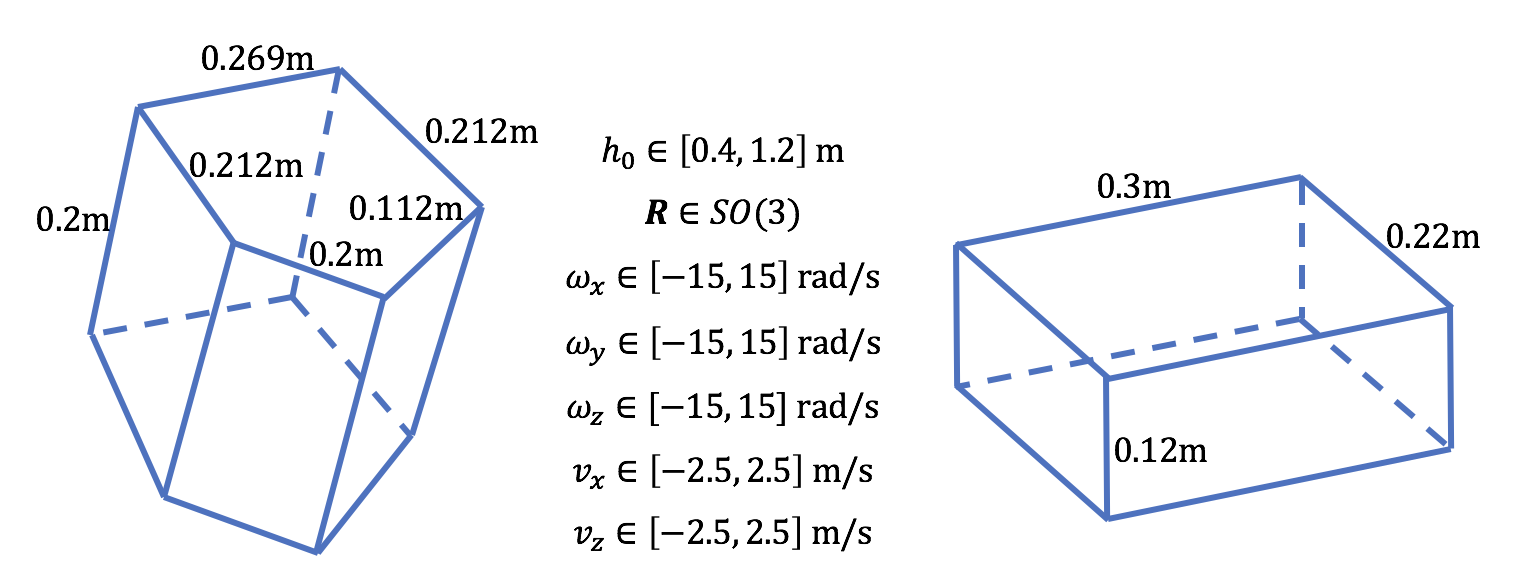}
  \caption{Range of initial states and the rigid bodies used in our experiments.}
  \label{fig:shapes}
\end{wrapfigure}


We evaluated our results against two baselines. The first baseline is the ground truth (GT), i.e. contact handling of DART. The second baseline is a purely data-driven (PDD) approach which learns regressors to directly predict impulse from the rigid body state, without using classifiers or solving analytical constraints. The input representation, network architecture and learning algorithm are the same between PDD and our regressors. However, we gave PDD $50\%$ more training data to reach comparable one-step accuracy (Fig. \ref{fig:params}). 


\subsection{Single step prediction}
We generated $40,000$ individual collisions to test each learned classifier and regressor. The accuracies are shown in Fig. \ref{fig:params}. Note that PDD can reach $90\%+$ accuracy when predicting individual collisions.

\subsection{Predicting trajectories of single rigid body}
We threw a box and a pentagon prism to the ground under gravity from various initial positions and velocities. The range of initial states and the geometries of the rigid bodies are detailed in Fig. \ref{fig:shapes}, where $h_0$, $\vc{R}$, $\omega$, $v$ indicate the ranges of the initial height, 3D orientation, angular velocity, and linear velocity respectively. Each simulated trajectory contains $800$ time steps, equivalent to $1.6$ second of motion. For each rigid body, we simulated $100$ trajectories with random initial states and reported three metrics: the average errors of the final horizontal distance, of the final orientation, and of the first collision impulse \footnote{The later collisions cannot be compared to the ground truth because the motions start to deviate after the first collision.} of each trajectory.

\begin{figure}
\centering
  \includegraphics[width=5.0in]{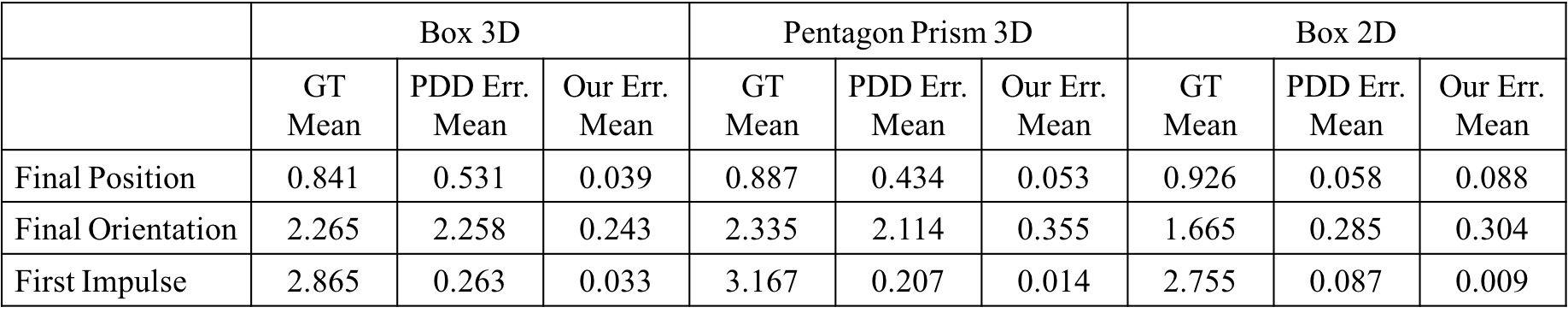}
  \caption{Results of the single body experiments.} 
  \label{fig:results}
\end{figure}

Fig. \ref{fig:results} shows the results in comparison with the two baselines, where for each metric, we show the mean of GT, the average error of PDD compared to GT, and the average error of our method compared to GT. In most cases, our method matches GT closely and is significantly better than PDD, demonstrating the advantages of using a classifier and analytical solutions. It is worth noting that both our method and PDD achieve similar accuracy in learning the regressors, but our method has much lower error in predicting the impulses. This is because when a collision instance is correctly classified as "static" or "detach", our method solves for an analytical solution which adds no error to the simulation. We also notice that PDD produces large errors in distance and orientation. This is because the small but persistent errors in impulse often result in perpetual movements instead of letting the rigid body come to rest. The erroneous behavior further highlights the advantage of identifying static cases and enforcing analytical constraints for those cases. For comparison, we also tested PDD and our method on a 2D rectangle. PDD performs much better on the planar throwing problem and is comparable to our method. This seems to suggest that PDD can only simulate consecutive bounces well when its regressor has a very high accuracy ($\approx 98\%$), which is much harder to achieve in 3D problems.

\begin{wrapfigure}{r}{4.0in}
\centering
  \includegraphics[width=4.0in]{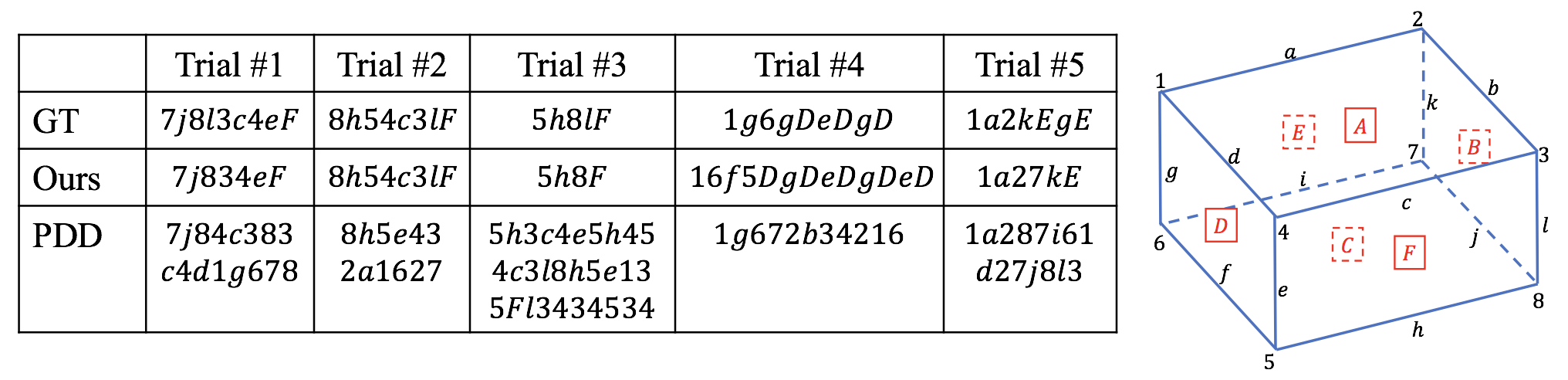}
  \caption{Contact event sequences of five random trials.}
  \label{fig:modes}
\end{wrapfigure}

Since 3D motions involve much more complex contact behaviors than 2D, we also compared the sequence of contact events (rolling, spinning etc.) in addition to the final state of the trajectory. Fig. \ref{fig:modes} shows the contact event sequences of the 3D box from five random throws. To represent a sequence, we used integers (1-8) to label the contacting vertices, lower case letters ($a$ to $l$) to label the contacting edges, and upper case letters ($A$ to $F$) to label contacting faces. The results show that our method produces similar contact events to GT while PDD produces wildly different contact events.

\subsection{Predicting trajectories of articulated rigid bodies}
We demonstrated our method on an articulated three link system connected by two revolute joints. The top of the first link is pinned to a fixed world space location. The chain started at a horizontal position and swang passively to the ground under gravity. We compared our method to ground truth and showed the motion sequences in the supplementary video. Though our method is only trained on the contact instances between an isolated distal body (\ie the third link) and the ground, we show that both contact impulses and joint constraint forces can be predicted or solved accurately. 




 \subsection{Anisotropic friction cone}

\begin{figure}
\centering
  \includegraphics[width=4.0in]{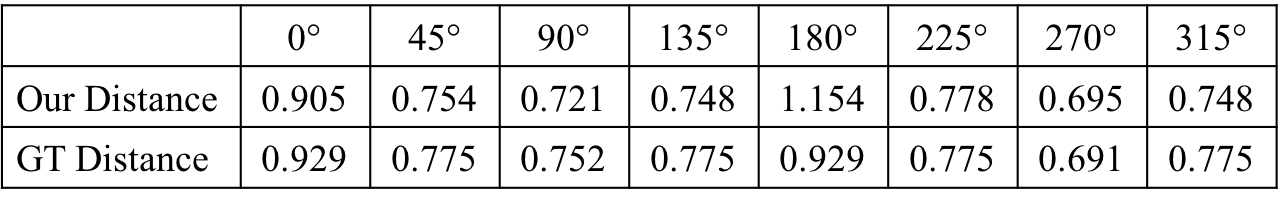}
  \caption{A box thrown in eight directions onto a surface with anisotropic materials.}
  \label{fig:unsym}
\end{figure}

We created a fictitious material which has an anisotropic friction cone. The friction coefficient of the ground is $1.5$ along z-axis and $0.75$ along x-axis. We collected training data from this simulated scenario and learned the classifiers and regressors using the same algorithms. During testing, we threw a box to the ground in eight directions. For each direction, we oriented the initial orientation and velocity to align with the throwing direction. Fig. \ref{fig:unsym} shows the distances traveled for each direction using our contact model (top) and using the GT simulator (bottom). The results show that our method is able to predict the outcome of collision for anisotropic materials.
 

\section{Conclusion and Limitations}
\label{sec:limitations}
We introduced a \emph{data-augmented} contact model that predicts the contact behaviors for a particular pair of near-rigid bodies or articulated rigid body systems. We evaluated our method on a set of 3D examples using simulated training data. The promising results indicate that our work could be a first step toward a contact model capable of predicting the 3D contact behaviors of real-world, near-rigid materials.

When collecting the training data from the real world, an isolated distal part of the robot will be used to create the collisions with the surface. The range of collision impulses should cover that of the anticipated collisions during the operation of the full robot. Although the data collection can be conducted in isolation without involving the entire robot, the data-efficiency remains a major concern. Since the dimension of the input and output space is relatively low, it is possible to use other function approximators, such as support vector machines or Gaussian processes, which might be more data-efficient than neural networks.

Our current algorithm assumes that one of the objects is stationary, and does not handle simultaneous contacts of multiple bodies, which limits its usage in many manipulation tasks. As immediate future directions, we plan to extend Algorithm \ref{algo_single_body} to two moving bodies by including states of both objects as input. Finally, combining our method with learnable contact detectors could further increase expressiveness of the model.

\newpage
\bibliography{references}

\begin{thebibliography}{37}
\providecommand{\natexlab}[1]{#1}
\providecommand{\url}[1]{\texttt{#1}}
\expandafter\ifx\csname urlstyle\endcsname\relax
  \providecommand{\doi}[1]{doi: #1}\else
  \providecommand{\doi}{doi: \begingroup \urlstyle{rm}\Url}\fi

\bibitem[Ajay et~al.(2018)Ajay, Wu, Fazeli, Bauza, Kaelbling, Tenenbaum, and
  Rodriguez]{ajay2018augmenting}
Anurag Ajay, Jiajun Wu, Nima Fazeli, Maria Bauza, Leslie~P Kaelbling, Joshua~B
  Tenenbaum, and Alberto Rodriguez.
\newblock Augmenting physical simulators with stochastic neural networks: Case
  study of planar pushing and bouncing.
\newblock \emph{arXiv preprint arXiv:1808.03246}, 2018.

\bibitem[Anitescu and Potra(2002)]{anitescu2002time}
Mihai Anitescu and Florian~A Potra.
\newblock A time-stepping method for stiff multibody dynamics with contact and
  friction.
\newblock \emph{International Journal for Numerical Methods in Engineering},
  55, 2002.

\bibitem[Baraff(1989)]{baraff1989analytical}
David Baraff.
\newblock Analytical methods for dynamic simulation of non-penetrating rigid
  bodies.
\newblock In \emph{ACM SIGGRAPH Computer Graphics}, volume~23, pages 223--232.
  ACM, 1989.

\bibitem[Baraff(1992)]{baraff1992dynamic}
David Baraff.
\newblock \emph{Dynamic Simulation of Non-penetrating Rigid Bodies}.
\newblock PhD thesis, Cornell University, 1992.

\bibitem[Bauza and Rodriguez(2017)]{bauza2017probabilistic}
Maria Bauza and Alberto Rodriguez.
\newblock A probabilistic data-driven model for planar pushing.
\newblock In \emph{Robotics and Automation (ICRA), 2017 IEEE International
  Conference on}, pages 3008--3015. IEEE, 2017.

\bibitem[Brogliato and Brogliato(1999)]{brogliato1999nonsmooth}
Bernard Brogliato and B~Brogliato.
\newblock \emph{Nonsmooth mechanics}.
\newblock Springer, 1999.

\bibitem[Byravan and Fox(2017)]{byravan2017se3}
Arunkumar Byravan and Dieter Fox.
\newblock Se3-nets: Learning rigid body motion using deep neural networks.
\newblock In \emph{Robotics and Automation (ICRA), 2017 IEEE International
  Conference on}, pages 173--180. IEEE, 2017.

\bibitem[Chang et~al.(2016)Chang, Ullman, Torralba, and
  Tenenbaum]{chang2016compositional}
Michael Chang, Tomer Ullman, Antonio Torralba, and Joshua~B Tenenbaum.
\newblock A compositional object-based approach to learning physical dynamics.
\newblock In \emph{Proceedings of the 5th International Conference on Learning
  Representations}, 2016.

\bibitem[Chebotar et~al.(2019)Chebotar, Handa, Makoviychuk, Macklin, Issac,
  Ratliff, and Fox]{chebotar2019closing}
Yevgen Chebotar, Ankur Handa, Viktor Makoviychuk, Miles Macklin, Jan Issac,
  Nathan Ratliff, and Dieter Fox.
\newblock Closing the sim-to-real loop: Adapting simulation randomization with
  real world experience.
\newblock In \emph{2019 International Conference on Robotics and Automation
  (ICRA)}, pages 8973--8979. IEEE, 2019.

\bibitem[Cottle and Dantzig(1968")]{COTTLE1968103}
Richard~W. Cottle and George~B. Dantzig.
\newblock Complementary pivot theory of mathematical programming.
\newblock \emph{Linear Algebra and its Applications}, 1\penalty0 (1):\penalty0
  103 -- 125, 1968".
\newblock ISSN 0024-3795.
\newblock \doi{https://doi.org/10.1016/0024-3795(68)90052-9}.
\newblock URL
  \url{http://www.sciencedirect.com/science/article/pii/0024379568900529}.

\bibitem[Degrave et~al.(2019)Degrave, Hermans, Dambre, and
  wyffels]{10.3389/fnbot.2019.00006}
Jonas Degrave, Michiel Hermans, Joni Dambre, and Francis wyffels.
\newblock A differentiable physics engine for deep learning in robotics.
\newblock \emph{Frontiers in Neurorobotics}, 13:\penalty0 6, 2019.
\newblock ISSN 1662-5218.
\newblock \doi{10.3389/fnbot.2019.00006}.
\newblock URL
  \url{https://www.frontiersin.org/article/10.3389/fnbot.2019.00006}.

\bibitem[Fazeli et~al.(2017)Fazeli, Zapolsky, Drumwright, and
  Rodriguez]{fazeli2017learning}
Nima Fazeli, Samuel Zapolsky, Evan Drumwright, and Alberto Rodriguez.
\newblock Learning data-efficient rigid-body contact models: Case study of
  planar impact.
\newblock In \emph{Conference on Robot Learning}, pages 388--397, 2017.

\bibitem[Fazeli et~al.(2020)Fazeli, Ajay, and Rodriguez]{fazeli2020long}
Nima Fazeli, Anurag Ajay, and Alberto Rodriguez.
\newblock Long-horizon prediction and uncertainty propagation with residual
  point contact learners.
\newblock In \emph{2020 IEEE International Conference on Robotics and
  Automation (ICRA)}, pages 7898--7904. IEEE, 2020.

\bibitem[Goyal et~al.(1989)Goyal, Ruina, and Papadopoulos]{goyal1989limit}
Suresh Goyal, Andy Ruina, and Jim Papadopoulos.
\newblock Limit surface and moment function descriptions of planar sliding.
\newblock In \emph{Robotics and Automation (ICRA), 1989 IEEE International
  Conference on}, pages 794--799. IEEE, 1989.

\bibitem[Heiden et~al.(2021)Heiden, Millard, Coumans, Sheng, and
  Sukhatme]{9560935}
Eric Heiden, David Millard, Erwin Coumans, Yizhou Sheng, and Gaurav~S.
  Sukhatme.
\newblock Neuralsim: Augmenting differentiable simulators with neural networks.
\newblock In \emph{2021 IEEE International Conference on Robotics and
  Automation (ICRA)}, pages 9474--9481, 2021.
\newblock \doi{10.1109/ICRA48506.2021.9560935}.

\bibitem[Hwangbo et~al.(2019)Hwangbo, Lee, Dosovitskiy, Bellicoso, Tsounis,
  Koltun, and Hutter]{hwangbo2019learning}
Jemin Hwangbo, Joonho Lee, Alexey Dosovitskiy, Dario Bellicoso, Vassilios
  Tsounis, Vladlen Koltun, and Marco Hutter.
\newblock Learning agile and dynamic motor skills for legged robots.
\newblock \emph{Science Robotics}, 4\penalty0 (26), 2019.

\bibitem[Jatavallabhula et~al.(2021)Jatavallabhula, Macklin, Golemo, Voleti,
  Petrini, Weiss, Considine, Parent-Levesque, Xie, Erleben, Paull, Shkurti,
  Nowrouzezahrai, and Fidler]{gradsim}
Krishna~Murthy Jatavallabhula, Miles Macklin, Florian Golemo, Vikram Voleti,
  Linda Petrini, Martin Weiss, Breandan Considine, Jerome Parent-Levesque,
  Kevin Xie, Kenny Erleben, Liam Paull, Florian Shkurti, Derek Nowrouzezahrai,
  and Sanja Fidler.
\newblock gradsim: Differentiable simulation for system identification and
  visuomotor control.
\newblock \emph{International Conference on Learning Representations (ICLR)},
  2021.
\newblock URL \url{https://openreview.net/forum?id=c_E8kFWfhp0}.

\bibitem[Jeong et~al.(2019)Jeong, Kay, Romano, Lampe, Rothorl, Abdolmaleki,
  Erez, Tassa, and Nori]{jeong2019modelling}
Rae Jeong, Jackie Kay, Francesco Romano, Thomas Lampe, Tom Rothorl, Abbas
  Abdolmaleki, Tom Erez, Yuval Tassa, and Francesco Nori.
\newblock Modelling generalized forces with reinforcement learning for
  sim-to-real transfer.
\newblock \emph{arXiv preprint arXiv:1910.09471}, 2019.

\bibitem[Jiang et~al.(2021)Jiang, Zhang, Ho, Bai, Liu, Levine, and
  Tan]{9561731}
Yifeng Jiang, Tingnan Zhang, Daniel Ho, Yunfei Bai, C.~Karen Liu, Sergey
  Levine, and Jie Tan.
\newblock Simgan: Hybrid simulator identification for domain adaptation via
  adversarial reinforcement learning.
\newblock In \emph{2021 IEEE International Conference on Robotics and
  Automation (ICRA)}, pages 2884--2890, 2021.
\newblock \doi{10.1109/ICRA48506.2021.9561731}.

\bibitem[Jourdan et~al.(1998)Jourdan, Alart, and Jean]{GS}
F.~Jourdan, P.~Alart, and M.~Jean.
\newblock A gauss-seidel like algorithm to solve frictional contact problems.
\newblock \emph{Computer Methods in Applied Mechanics and Engineering}, 155,
  1998.

\bibitem[Kloss et~al.(2018)Kloss, Schaal, and Bohg]{kloss2018}
Alina Kloss, Stefan Schaal, and Jeannette Bohg.
\newblock Combining learned and analytical models for predicting action
  effects.
\newblock \emph{arXiv:1710.04102}, 2018.

\bibitem[Le~Lidec et~al.(2021)Le~Lidec, Kalevatykh, Laptev, Schmid, and
  Carpentier]{9363565}
Quentin Le~Lidec, Igor Kalevatykh, Ivan Laptev, Cordelia Schmid, and Justin
  Carpentier.
\newblock Differentiable simulation for physical system identification.
\newblock \emph{IEEE Robotics and Automation Letters}, 6\penalty0 (2):\penalty0
  3413--3420, 2021.
\newblock \doi{10.1109/LRA.2021.3062323}.

\bibitem[Lee et~al.(2018)Lee, Grey, Ha, Kunz, Jain, Ye, Srinivasa, Stilman, and
  Liu]{DART}
Jeongseok Lee, Michael~X. Grey, Sehoon Ha, Tobias Kunz, Sumit Jain, Yuting Ye,
  Siddhartha~S. Srinivasa, Mike Stilman, and C.~Karen Liu.
\newblock {DART}: Dynamic animation and robotics toolkit.
\newblock \emph{The Journal of Open Source Software}, 2018.
\newblock \doi{10.21105/joss.00500}.
\newblock URL \url{https://doi.org/10.21105/joss.00500}.

\bibitem[Lemke and Howson(1964)]{Lemke}
C.~Lemke and J.~Howson.
\newblock Equilibrium points of bimatrix games.
\newblock \emph{SIAM Journal on Applied Mathematics}, 12, 1964.

\bibitem[Lerer et~al.(2016)Lerer, Gross, and Fergus]{lerer2016learning}
Adam Lerer, Sam Gross, and Rob Fergus.
\newblock Learning physical intuition of block towers by example.
\newblock In \emph{International Conference on Machine Learning}, pages
  430--438, 2016.

\bibitem[Oden and Martins(1985)]{oden1985models}
JT~Oden and JAC Martins.
\newblock Models and computational methods for dynamic friction phenomena.
\newblock \emph{Computer methods in applied mechanics and engineering}, 52,
  1985.

\bibitem[Olsson et~al.(1998)Olsson, {\AA}str{\"o}m, De~Wit, G{\"a}fvert, and
  Lischinsky]{olsson1998friction}
Henrik Olsson, Karl~Johan {\AA}str{\"o}m, Carlos~Canudas De~Wit, Magnus
  G{\"a}fvert, and Pablo Lischinsky.
\newblock Friction models and friction compensation.
\newblock \emph{Eur. J. Control}, 4\penalty0 (3):\penalty0 176--195, 1998.

\bibitem[Pfrommer et~al.(2020)Pfrommer, Halm, and
  Posa]{pfrommer2020contactnets}
Samuel Pfrommer, Mathew Halm, and Michael Posa.
\newblock Contactnets: Learning discontinuous contact dynamics with smooth,
  implicit representations.
\newblock \emph{arXiv preprint arXiv:2009.11193}, 2020.

\bibitem[Pizzuto and Mistry(2021)]{pizzuto2021physics}
Gabriella Pizzuto and Michael Mistry.
\newblock Physics-penalised regularisation for learning dynamics models with
  contact.
\newblock In \emph{Learning for Dynamics and Control}, pages 611--622. PMLR,
  2021.

\bibitem[Popova and Popov(2015)]{Elena}
Elena Popova and Valentin~L. Popov.
\newblock The research works of {C}oulomb and {A}montons and generalized laws
  of friction.
\newblock \emph{Friction}, 3, 2015.

\bibitem[Powell(1970)]{powell1970hybrid}
Michael~JD Powell.
\newblock A hybrid method for nonlinear equations.
\newblock \emph{Numerical methods for nonlinear algebraic equations}, 1970.

\bibitem[Rao et~al.(2020)Rao, Harris, Irpan, Levine, Ibarz, and
  Khansari]{rao2020rl}
Kanishka Rao, Chris Harris, Alex Irpan, Sergey Levine, Julian Ibarz, and Mohi
  Khansari.
\newblock Rl-cyclegan: Reinforcement learning aware simulation-to-real.
\newblock In \emph{Proceedings of the IEEE/CVF Conference on Computer Vision
  and Pattern Recognition}, pages 11157--11166, 2020.

\bibitem[Stewart and Trinkle(2000)]{stewart2000implicit}
David Stewart and Jeffrey~C Trinkle.
\newblock An implicit time-stepping scheme for rigid body dynamics with
  {C}oulomb friction.
\newblock In \emph{Robotics and Automation (ICRA), 2000 IEEE International
  Conference on}, volume~1, pages 162--169. IEEE, 2000.

\bibitem[Strecke and Stueckler(2021)]{strecke2021_diffsdfsim}
Michael Strecke and Joerg Stueckler.
\newblock {DiffSDFSim}: Differentiable rigid-body dynamics with implicit
  shapes.
\newblock In \emph{International Conference on {3D} Vision ({3DV})}, December
  2021.

\bibitem[Todorov(2014)]{todorov2014convex}
Emanuel Todorov.
\newblock Convex and analytically-invertible dynamics with contacts and
  constraints: Theory and implementation in {M}u{J}o{C}o.
\newblock \emph{Robotics and Automation (ICRA), 2014 IEEE International
  Conference on}, pages 6054--6061, 2014.

\bibitem[Yu et~al.(2016)Yu, Bauza, Fazeli, and Rodriguez]{yu2016more}
Kuan-Ting Yu, Maria Bauza, Nima Fazeli, and Alberto Rodriguez.
\newblock More than a million ways to be pushed. a high-fidelity experimental
  dataset of planar pushing.
\newblock In \emph{Intelligent Robots and Systems (IROS), 2016 IEEE/RSJ
  International Conference on}, pages 30--37. IEEE, 2016.

\bibitem[Zhou et~al.(2016)Zhou, Paolini, Bagnell, and Mason]{zhou2016convex}
Jiaji Zhou, Robert Paolini, J~Andrew Bagnell, and Matthew~T Mason.
\newblock A convex polynomial force-motion model for planar sliding:
  Identification and application.
\newblock In \emph{Robotics and Automation (ICRA), 2016 IEEE International
  Conference on}, pages 372--377. IEEE, 2016.

\end{thebibliography}

\end{document}